\documentclass[letterpaper, 10 pt, conference]{ieeeconf}  

\IEEEoverridecommandlockouts         
\usepackage{graphicx} 
\usepackage[utf8]{inputenc}
\usepackage{cite}
\usepackage{url}
\usepackage{hyperref}
\usepackage{color, colortbl}
\usepackage{xspace}
\usepackage{algorithm}
\usepackage[noend]{algpseudocode}
\usepackage{siunitx}
\usepackage{booktabs}
\usepackage{multirow}
\usepackage{fancyhdr}

\usepackage{mathabx}
\makeatletter
\newcommand\incircbin
{%
  \mathpalette\@incircbin
}
\newcommand\@incircbin[2]
{%
  \mathbin%
  {%
    \ooalign{\hidewidth$#1#2$\hidewidth\crcr$#1\ovoid$}%
  }%
}
\DeclareRobustCommand*{\oeq}{\incircbin{=}}
\makeatother

\newcommand{\stt}[1]{{\small\texttt{#1}}}

\title{\LARGE \bf Deliberation in autonomous robotic surgery: a framework for handling anatomical uncertainty }

\author{Eleonora Tagliabue$^*$, Daniele Meli$^*$, Diego Dall'Alba, Paolo Fiorini%
\thanks{This project has received funding from the European Research Council (ERC) under the European Union's Horizon 2020 research and innovation programme under grant agreement No. 742671 (ARS).}
\thanks{*Authors contributed equally to the work.}
\thanks{Authors are with Dept. of Computer Science, University of Verona, Italy.
Contact author: Daniele Meli, \texttt{daniele.meli@univr.it}}
}

\begin{document}

\maketitle
\thispagestyle{empty}
\pagestyle{empty}

\pagestyle{empty}

\thispagestyle{fancy}
\renewcommand{\headrulewidth}{0pt}
\renewcommand{\footrulewidth}{0.5pt}
\cfoot{\footnotesize{\copyright 2022 IEEE. Personal use of this material is permitted. Permission from IEEE must be obtained for all other uses, in any current or future media, including reprinting/republishing this material for advertising or promotional purposes, creating new collective works, for resale or redistribution to servers or lists, or reuse of any copyrighted component of this work in other works.}}

\begin{abstract}
Autonomous robotic surgery requires deliberation, i.e. the ability to plan and execute a task adapting to uncertain and dynamic environments. Uncertainty in the surgical domain is mainly related to the partial pre-operative knowledge about patient-specific anatomical properties. In this paper, we introduce a logic-based framework for surgical tasks with deliberative functions of monitoring and learning. The DEliberative Framework for Robot-Assisted Surgery (DEFRAS) estimates a pre-operative patient-specific plan, and executes it while continuously measuring the applied force obtained from a biomechanical pre-operative model. Monitoring module compares this model with the actual situation reconstructed from sensors. In case of significant mismatch, the learning module is invoked to update the model, thus improving the estimate of the exerted force. DEFRAS is validated both in simulated and real environment with da Vinci Research Kit executing soft tissue retraction. Compared with state-of-the-art related works, the success rate of the task is improved while minimizing the interaction with the tissue to prevent unintentional damage.
\end{abstract}


\section{INTRODUCTION}
In recent years, research in surgical robotics has focused on the automation of surgical tasks, for its potential to provide physical and cognitive support to the surgeon \cite{yang2017medical}.
Existing works in this field have mainly considered structured and controlled environments \cite{d2021accelerating}.
This assumption makes current approaches still far from the applicability to real clinical scenarios, which consist of deformable tissues whose mechanical properties are unknown pre-operatively, leading to intra-operative uncertainty \cite{attanasio2021autonomy}.

To cope with real deformable anatomical environments, an Autonomous Robotic Surgery (ARS) system shall integrate the following skills:
\begin{itemize}
    \item it should have \textit{deliberative} capabilities \cite{ingrand2017deliberation}, i.e. \emph{monitoring} the environment through \emph{sensors} while \emph{acting}, in order to \emph{reason} and \emph{plan} a new strategy when the expected behavior does not match reality. Moreover, it must adapt its prior model by \emph{learning} from observations, to minimize uncertainty and critical events;
    \item it should be \emph{informed} from pre-operative anatomical data and task-level expert knowledge, in order to devise a patient-specific strategy for the intervention;
    \item it should provide \emph{interpretable} plans, which can be easily understood by a supervising surgeon, to guarantee reliability of the ARS system \cite{fiazza2021design,attanasio2021autonomy}, as required from level 2 of autonomy \cite{yang2017medical}.
\end{itemize}

\begin{figure}[t]
	\centering
	\includegraphics[width=.89\linewidth]{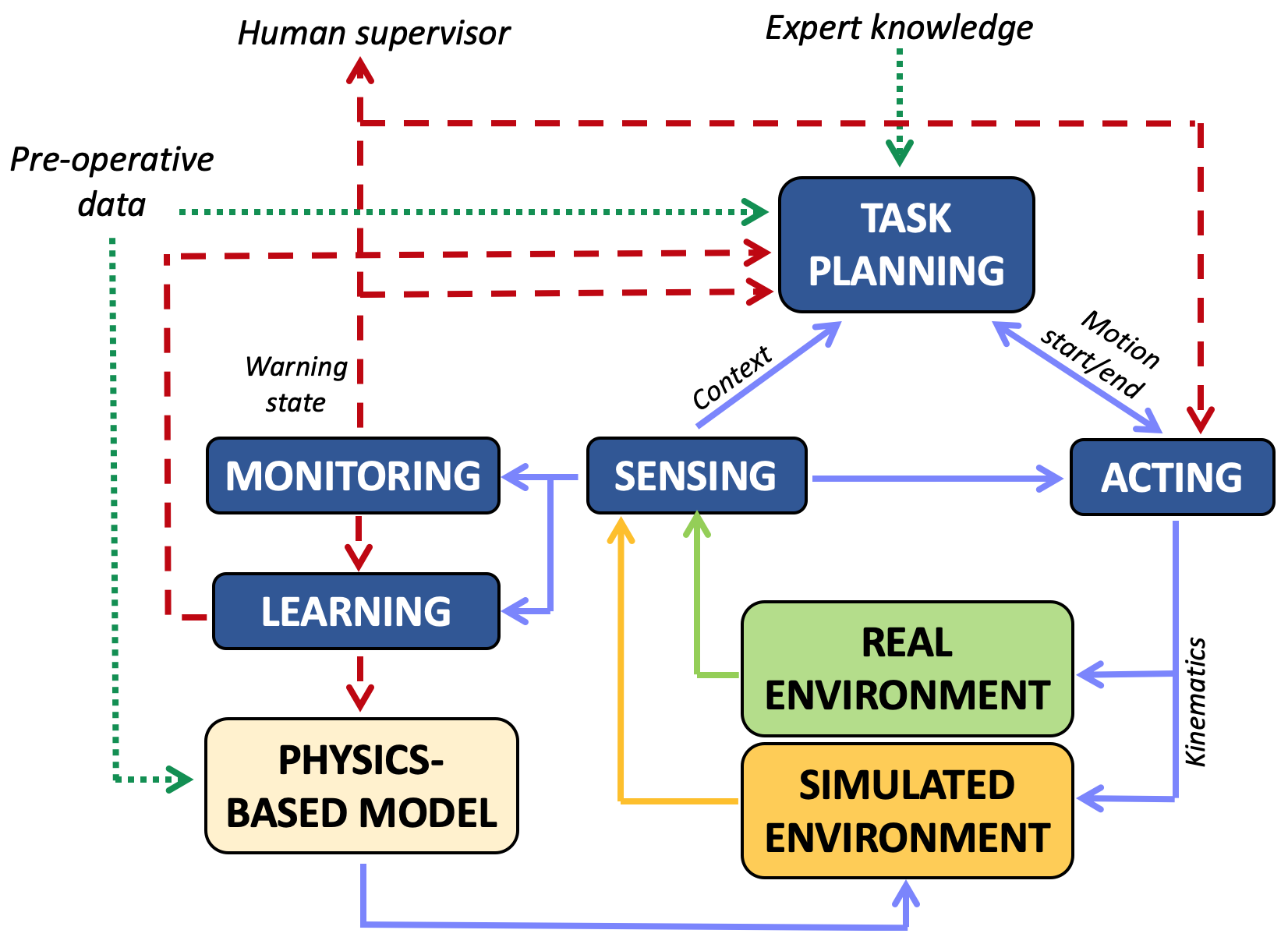}
	\caption{DEFRAS architecture (deliberative functions in blue boxes).
 	 Green lines represent pre-operative information. Red lines trigger actions only when monitoring threshold is exceeded.}
	\label{fig:framework}
\end{figure}

To address autonomy in real deformable surgical environments, we rely on a biomechanical simulation initialized with pre-operative information both to generate a patient-specific surgical plan \emph{pre-operatively} and to detect possible unexpected behavior \emph{intra-operatively}, via comparison with real sensor data.
The specific contributions of this work are:
\begin{enumerate}
	\item we present DEFRAS, a DEliberative Framework for Robot-Assisted Surgery that integrates skills necessary to cope with the uncertainties and the dynamic behavior of anatomical environments, while guaranteeing task completion and respect of safety constraints;
	\item we demonstrate it for a representative surgical task with uncertainty, i.e. soft tissue retraction, both in simulated and real experiments with the da Vinci Research Kit (dVRK) \cite{kazanzides2014open};
	\item we publicly share the framework at \url{https://gitlab.com/altairLab/tissue_retraction.git}.
\end{enumerate}

The paper is organized as follows. In Section \ref{sec:sota} we review recent frameworks for surgical robotic autonomy under uncertainty and generic frameworks for deliberation; we then describe DEFRAS and our implementation of the tissue retraction task in Sections~\ref{sec:framework} and \ref{sec:tr_defras} respectively; Section~\ref{sec:exp} describes the conducted validation experiments and Section~\ref{sec:res} reports the obtained results. 
Finally, we draw our conclusions in Section~\ref{sec:conclusion}.
\section{RELATED WORKS}
\label{sec:sota}
Uncertainty poses challenges to ARS systems relying only on prior knowledge about anatomical models and/or task description \cite{nagy2019dvrk}. 
Previous works have tried to deal with the different sources of uncertainty in the surgical domain, e.g. implementing strategies for accurate instrument and anatomy localization \cite{sen2016automating, shademan2016supervised}, online update of soft tissues models \cite{wu2020leveraging, liu2021real}, workflow recognition \cite{de2019cognitive} and task model refinement \cite{Hubot, murali2015learning}.
Model predictive control has also been exploited to deal with uncertainty at motion planning level \cite{de2019cognitive,kehoe2014autonomous}. 

While existing works have attempted either to reduce uncertainty at sensing level or to deal with it at motion/task level, in this paper we address both aspects by introducing deliberative functions of autonomous agents \cite{ingrand2017deliberation}. In particular, we integrate specific modules for uncertainty \emph{monitoring} and pre-operative model refinement through online \emph{learning} within a framework for robot-assisted surgery \cite{meli2021autonomous} that guarantees task-level interpretability with logic programming and supports online re-planning. 
Several paradigms for deliberation have been proposed in the field of artificial intelligence, focusing on different aspects of autonomy, e.g. planning \cite{Bratman1987-BRAIPA, georgeff1987reactive, bratman1988plans} and learning
\cite{bolisani2017knowledge}.
However, in order to guarantee adaptability to diverse complex tasks, here we focus on the integration and parallel development of different skills for autonomy, which can be improved individually depending on the task of interest.
For this reason, our framework builds on the modular architecture and definitions of deliberative functions described in \cite{ingrand2017deliberation}, which reports a large number of robotic applications.

\section{DEFRAS: DELIBERATIVE FRAMEWORK FOR ROBOT-ASSISTED SURGERY}
\label{sec:framework}
Our deliberative architecture is shown in Fig.~\ref{fig:framework}. 
A \emph{task planning} module computes the task plan, i.e. the sequence of actions to solve the task.
Actions are elementary operations which correspond to a single motion trajectory or joint-level command (e.g., reaching a target or grasping), according to the granularity definitions described in \cite{Lalys}.
The plan is computed based on available expert procedural knowledge, conveniently expressed in the declarative logic programming formalism of Answer Set Programming (ASP) \cite{lifschitz2002answer, calimeri2020asp}. 
The ASP task description encodes entities, i.e. definitions of actions, tools and relevant anatomical features as logic variables and predicates (\emph{atoms}).
Logical implications between atoms are implemented to define pre-conditions and effects of actions, and constraints (e.g., infeasibility of some actions).
\emph{Answer sets} are then computed by ASP solver Clingo \cite{gebser2008user} as the minimal sets of grounded atoms (i.e., with assignments of variables), hence including the shortest plans along with the corresponding environmental features. This enhances task-level interpretability. 
Throughout task execution, the \emph{sensing} module analyses the state of the robot and the environment, to provide high-level environmental features for task planning and targets for motion planning (\emph{acting} module). 
Moreover, it recognizes whether critical conditions occur, in which cases it triggers task re-planning or motion interruption to prevent risks.
The \emph{monitoring} module continuously estimates if the internal model of the environment, given by a patient-specific simulation initialized with pre-operative information and running in parallel, is coherently representing the actual scenario.
To this aim, a task-specific metric assessing the level of discrepancy between the real and the simulated environments is computed starting from sensing information. Depending on discrepancy, different \emph{warning states} are raised, associated  with specific strategies at task-motion level to increase reliability and accommodate safety.
Finally, the \emph{learning} module is in charge of updating the pre-operative model from real data in warning states, to compensate for environmental uncertainty caused by inaccurate model parametrization.

\section{TISSUE RETRACTION WITH DEFRAS}
\label{sec:tr_defras}
DEFRAS functionalities are tested on autonomous execution of Tissue Retraction (TR), a representative surgical task which involves the interaction with soft anatomical tissues, whose properties might be not fully known before the intervention \cite{pore2021learning,pore2021saferl}.
It consists in grasping and retracting soft tissues to expose a hidden Region Of Interest (ROI) to an intra-operative camera (Fig.~\ref{fig:config-real}). The tissue is attached to surrounding anatomies in correspondence of Attachment Points (APs). 
Previous works for autonomous TR have proposed either to generate the plan by optimizing some metrics \cite{patil2010toward,jansen2009surgical,attanasio2020autonomous,tagliabue2020soft,pore2021saferl} or to learn the task from human demonstrations \cite{shin2019autonomous,pedram2019toward,pore2021learning}. However, none of them integrates methods to reason on the current environment and deal with unexpected situations which might occur in real surgery. 

\subsection{TR implementation with DEFRAS}
\label{subsec:tr_defras}
The task is executed on the dVRK, whose Patient-Side Manipulators (PSMs) are equipped with ProGrasp\textregistered\, surgical grasping tools. In general, ROI exposure is obtained primarily by pulling up the tissue after grasping. The grasping location on the tissue can be reached by either one of the PSMs, depending on \emph{reachability} conditions defined with respect of the tissue center\footnote{With reference to Fig.~\ref{fig:config-real}, a point on the tissue is reachable by a PSM if it lies on the same side as the PSM with respect to the $x-z$ plane.}, to avoid instrument collision during operation. However, pulling is not always successful due to kinematic limitations imposed by the anatomical environment and the insertion points of the instruments in the body, which define fixed pivots for the robotic arms.
Furthermore, pulling may induce high forces which can damage the tissue. 
In these cases, alternative actions can be performed, such as moving the tissue in a parallel plane to the tissue surface, either folding it away from the intra-operative camera or possibly towards the ROI.

In this work, we create a physics-based model of the anatomy leveraging on tissue geometric and mechanical properties as well as locations of the ROI and Attachment Points (APs) obtained from pre-operative data. Such model is used to initialize a simulation which is exploited to find a patient-specific task plan in the pre-operative phase. We select the plan that maximizes the economy of motion and reduces the risk of critical events (see Section~\ref{sec:exp} for details).
The pre-operative plan is then executed in the real environment. 

During execution, the physics-based simulation is run in parallel, and the state of the real environment is continuously monitored with intra-operative sensors (Fig.~\ref{fig:framework}). When mismatch between simulation and reality is identified, the pre-operative plan may become sub-optimal or even infeasible, hence simulation parameters may be updated and specific strategies to improve the plan are implemented. In this work, we focus on the update of APs. Although APs can be initialized pre-operatively based on anatomical knowledge, previous works have shown that their proper definition is possible only during interaction with the tissue \cite{nikolaev2020estimation,tagliabue2021intra}. The reason why we focus on APs is twofold. Firstly, they constitute model boundary conditions, thus represent an important parameter influencing simulation accuracy \cite{nikolaev2020estimation,tagliabue2021intra}. Secondly, knowledge of APs location is very important in TR to prevent the pulling of highly constrained areas, minimizing the risk of tissue damage.
The online simulation also continuously provides an estimate of the applied force on the tissue, compensating for the lack of force feedback in surgical robotic systems. 

\subsection{Task planning}
\label{subsec:framework_planning}
Possible actions implemented in ASP task description are \emph{reaching a grasping point, opening gripper, grasping, pulling} the tissue and folding it, either \emph{moving planarly towards the ROI} or \emph{moving away from the intra-operative camera}. They are expressed as predicates in plain English \emph{action(tool, object, property, time)}, e.g., \stt{reach(A, point, (X,Y), t)}, being \stt{A} a generic tool, \stt{(X,Y)} the coordinates of a grasping point and \stt{t} a discrete time step.
Possible grasping points are defined as ASP entities to be evenly spaced on the tissue surface.
This allows to exploit built-in ASP optimization to select the most reasonable grasping point in case of re-planning, as the one farthest from APs and closest to ROI \cite{meli2021autonomous}. This is a commonsense approach to minimize tissue damage and expose ROI fast, but different strategies are also possible, e.g., \cite{jansen2009surgical}.
Relevant environmental features include \emph{reachability} conditions, \emph{mutual locations of APs, ROI and tools} and conditions on \emph{force} and \emph{height} of PSMs, as in \cite{meli2021autonomous}. They are expressed as predicates similarly to actions (e.g., \stt{max\_force(A, t)}).
Logical implications encode action pre-conditions, effects, constraints and the goal (i.e., the ROI must be eventually exposed). For instance, \emph{grasp only reachable points} becomes \stt{grasp(A, point, (X,Y), t) :- reachable(A, point, (X,Y), t)}, where \stt{:-} is ASP notation for $\leftarrow$.

\subsection{Sensing and acting}
The sensing module continuously receives the state of the robot and the environment.
The state of the robot consists of the position of tools and opening angles of grippers.
The state of the environment includes the point cloud of tissue surface $PC_t$ and the ROI from the intra-operative camera; 
he position of the ROI, known APs and estimated forces at each tissue point from the simulation. 
Geometric reasoning is performed in real time on input data to ground anatomical features for task planning and anomaly detection. 
In particular, the sensing module continuously checks whether excessive force is applied to the tissue. When the maximum tissue force $F$ exceeds a pre-defined threshold $\epsilon$, DEFRAS first tries to reduce tissue solicitation by decreasing motion velocity \cite{wang2017influence}, otherwise motion is interrupted and re-planning is triggered.
Input data is also used to compute target positions for actions (e.g., locations of grasping points for \emph{reach}). Orientations of PSMs are kept constant for simplicity.
The acting module then translates ASP actions to joint (\emph{grasp, release}) or Cartesian trajectories for the dVRK.  

\subsection{Monitoring}
\label{subsec:framework_monitoring}
The monitoring module continuously receives $PC_t$ and the point cloud of the simulated tissue $PC_s$ from the sensing module, and computes a discrepancy metric $\mu$, chosen as the median Euclidean distance of closest points between $PC_t$ and $PC_s$ (Alg.~\ref{alg:monitoring}). The distance is normalized with respect to its rest value (i.e., before grasping occurs) to compensate for offsets induced by registration error and sensor noise.
Inspired from recent works as \cite{machin2016smof}, the monitoring module considers 3 thresholds $\delta_1, \delta_2, \delta_3$ (empirically set to 2, 4 and 6\,mm) to classify the system state.
n particular, when $\mu > \delta_1$ the motion velocity is reduced for caution, to minimize the risk of applying excessive forces \cite{wang2017influence} and allow prompt response to anomalies.
If $\delta_2$ is exceeded, motion is interrupted and re-planning is asked at task level. 
In both states, the learning module is triggered to update environmental knowledge exploiting data from sensors.
Overcoming $\delta_3$ results in interruption of DEFRAS and human intervention is required.

\begin{algorithm}[t]
    \caption{Monitoring module}
    \label{alg:monitoring}
    \begin{algorithmic}[1]
        \State \textbf{Input}: $PC_t$, $PC_s$, motion rate $r$, \stt{grasped\_tissue}
        \State \textbf{Output}: Warning state $s$, motion rate $r$, trigger learning \stt{learn}
        \State \textbf{Init}: $d_{rest} = []$
        \While{not \stt{task\_ended}}
        \While{not \stt{grasped\_tissue}}
            \State $d_{rest}$.append($||PC_t - PC_s||_{2}$)
        \EndWhile
            \State $d_{rest} = $mean$(d_{rest})$
            \State $d = ||PC_t - PC_s||_{2} $
            \State $\mu = d - d_{rest}$
            \If{$\delta_1 < \mu \leq \delta_2 $}
                \State \Return $s=1; r=\frac{r}{2};$ \stt{learn=True}
            \ElsIf{$\delta_2 < \mu \leq \delta_3 $}
                \State \Return $s=2; r=0;$ \stt{learn=True}
            \ElsIf{$\mu > \delta_3 $}
                 \State \Return $s=3; r=0;$ \stt{learn=False}
            \Else
                \State \Return $s=0; r=r;$ \stt{learn=False}
            \EndIf
        \EndWhile
    \end{algorithmic}
\end{algorithm}

\subsection{Learning}
\label{subsec:framework_learning}
The learning module focuses on the intra-operative update of APs whenever required by the monitoring module. 
Update relies on BANet, a deep neural network which predicts APs starting from the tissue pre-operative geometry and the current observed $PC_t$ \cite{tagliabue2021data}. 
This method has proved suitable for intra-operative model refinement, requiring a very low inference time and even improving biomechanical model accuracy \cite{tagliabue2021intra}. 
The new estimate of APs is used both to refine the simulated physics-based model, thus improving its reliability (especially of the force estimate), and to update the knowledge of the regions to avoid in case of re-planning. 

\section{EXPERIMENTS}
\label{sec:exp}
\begin{figure*}[th]
\includegraphics[width=.99\linewidth]{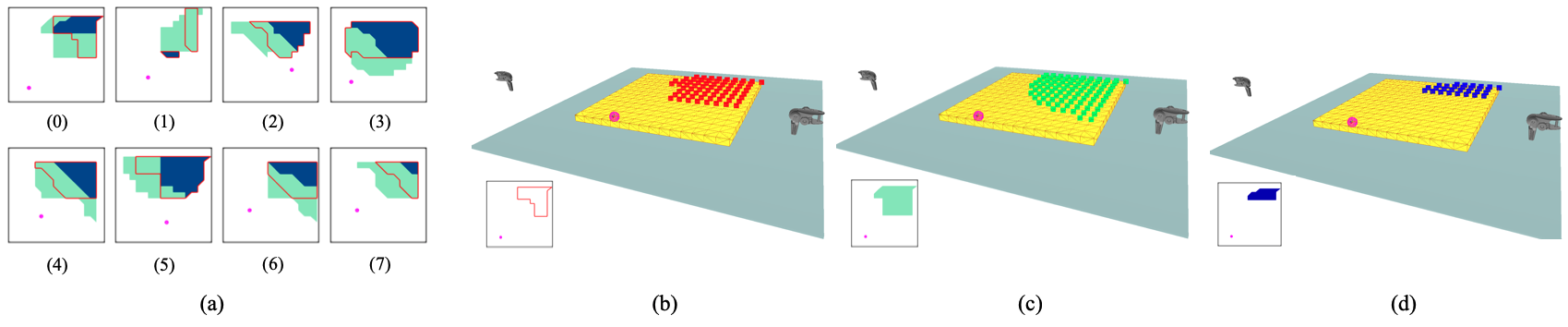}
\caption{(a) Configurations of APs used for the experiments in simulation. The red outline delimits the pre-operative configuration used to generate the plan; considered actual simulations are in blue ($\ominus$) and green ($\oplus$); magenta sphere represents the ROI. (b) Zoom in on the pre-operative configuration for case 0; (c) Zoom in on the $\oplus$ actual configuration for case 0; (d) Zoom in on the $\ominus$ configuration for case 0.}
\label{fig:config-sim}
\end{figure*}

We consider the problem of retracting a thin layer of soft tissue (120x120x5\,mm) to expose at least 80\% of ROI surface to the endoscopic camera, which is placed at the main surgical viewpoint (Fig.~\ref{fig:config-real}). 
In particular, we aim at assessing the capability of DEFRAS to accomplish TR task, coping with inaccurate parametrization of the available model of the environment from pre-operative information.

For monitoring purposes, we continuously capture the tissue point cloud $PC_t$ with an additional Intel\textregistered \ Realsense D435 RGBD sensor, placed opposite to the endoscope to maximize tissue visibility during manipulation, as done in \cite{tagliabue2021intra}. 
The RGBD sensor, endoscopic camera and the PSMs are calibrated with respect to a common reference frame placed at the center of the tissue, using the methodology described in \cite{roberti2020improving}. 
The simulated environment is aligned to the same reference frame and a simulated RGBD sensor placed in the same position as the real one provides the point cloud of the simulated tissue $PC_s$.
Tissue deformable behavior is modelled relying on the laws of continuum mechanics, solved with the finite element method using the SOFA framework \cite{faure2012sofa}. The tissue is described as a linear elastic material with Young's modulus $3$\,kPa and Poisson ratio $0.45$ \cite{tagliabue2021intra}, relying on the corotational formulation to handle large deformations while guaranteeing computational efficiency. 
For all experiments in this paper, we consider a set of 25 possible grasping points, regularly spaced over a 5x5 grid to maximize computational efficiency \cite{meli2021autonomous}.
We design 17 pre-operative configurations of tissue with different locations for APs and ROI (Fig.~\ref{fig:config-sim}a-\ref{fig:config-real}b).

As an initial step, we perform an exploratory search to identify a reasonable value for the force threshold $\epsilon$.
For each pre-operative configuration we launch DEFRAS only in simulation, deactivating learning and monitoring and ignoring force measurements. 
We compute plans for all possible grasping points under these conditions, and we record the estimated forces from the simulation. 
We obtain a median force $F_m =$ \SI{0.15}{\N} among all 425 executions. $F_m$ is chosen to represent the typical force on the tissue in the considered setup, hence we set $\epsilon = 2F_m =$ \SI{0.3}{\N}.

Afterwards, for each pre-operative configuration we generate an optimal patient-specific TR plan as follows.
We launch DEFRAS in simulation without learning and monitoring modules, but now including force measurements. 
We record plans originated picking all possible grasping points, and we select only plans maximizing the exposure goal, i.e. reaching a minimum visibility of 80\%. Among these, we select the 3 plans minimizing $F_{max}$, which is the maximum recorded tissue force over the entire execution. Finally, we choose the optimal plan as the one with the lowest number of actions, thus minimizing interaction with tissue and obtaining the maximum economy of motion.
We validate the full DEFRAS both in simulation (8 out of 17 pre-operative configurations) and on real setup (9 out of 17 pre-operative configurations).

\subsection{Simulation Experiments}
\label{subsec:sim_exp}
Each of the 8 pre-operative scenarios considered in our simulation experiments is characterized by a different configuration of APs (red outlines in Fig.~\ref{fig:config-sim}) extracted as proposed in \cite{tagliabue2021data}. 
For each pre-operative scenario, we generate optimal plans following the strategy described above.
Each pre-operative plan is executed by DEFRAS on two validation scenarios: the former with fewer APs (between 20-60\%) and the latter with more APs (between 140-160\%) with respect to the corresponding pre-operative configuration. 
These experiments emulate the realistic situation where initial knowledge of APs is not precise, thus the pre-operative plan might need adjustment due to mismatch between the current and the simulated environments. 
To run these experiments, we rely on two parallel simulations: the former (\emph{pre-operative simulation}) represents the simulated environment in the scheme of Fig.~\ref{fig:framework}, estimating the tissue forces and initialized with pre-operative APs; the latter (\emph{actual simulation}) represents the validation scenario with either fewer or more APs than the pre-operative simulation, with simulated dVRK and cameras. 
Visibility in the actual simulation is computed by counting the number of visible ROI points from the simulated endoscope, assessed via ray-casting technique. 
The monitoring module continuously compares the tissue point clouds from the two simulations, triggering specific interventions according to Alg.~\ref{alg:monitoring}. To assess the performance of the monitoring module, we compute the success rate of APs update with BANet as ratio between the number of times the update results in $\mu \leq \delta_1$ in Alg.~\ref{alg:monitoring} (no warning from monitoring) and the number of times BANet is triggered because $\mu > \delta_1$ (warning state). 

We compare the performance of DEFRAS with respect to FRAS \cite{meli2021autonomous}, the most similar framework in the surgical literature, in order to show the advantages of monitoring and learning modules in a deliberative architecture for autonomous robotic surgery. 
Therefore, we let FRAS perform autonomous TR on the actual simulations. It is worth recalling that FRAS does not rely on any pre-operative plan. 

\subsection{Real World Experiments}
\label{subsec:real_exp}
\begin{figure}[b]
\includegraphics[width=.99\linewidth]{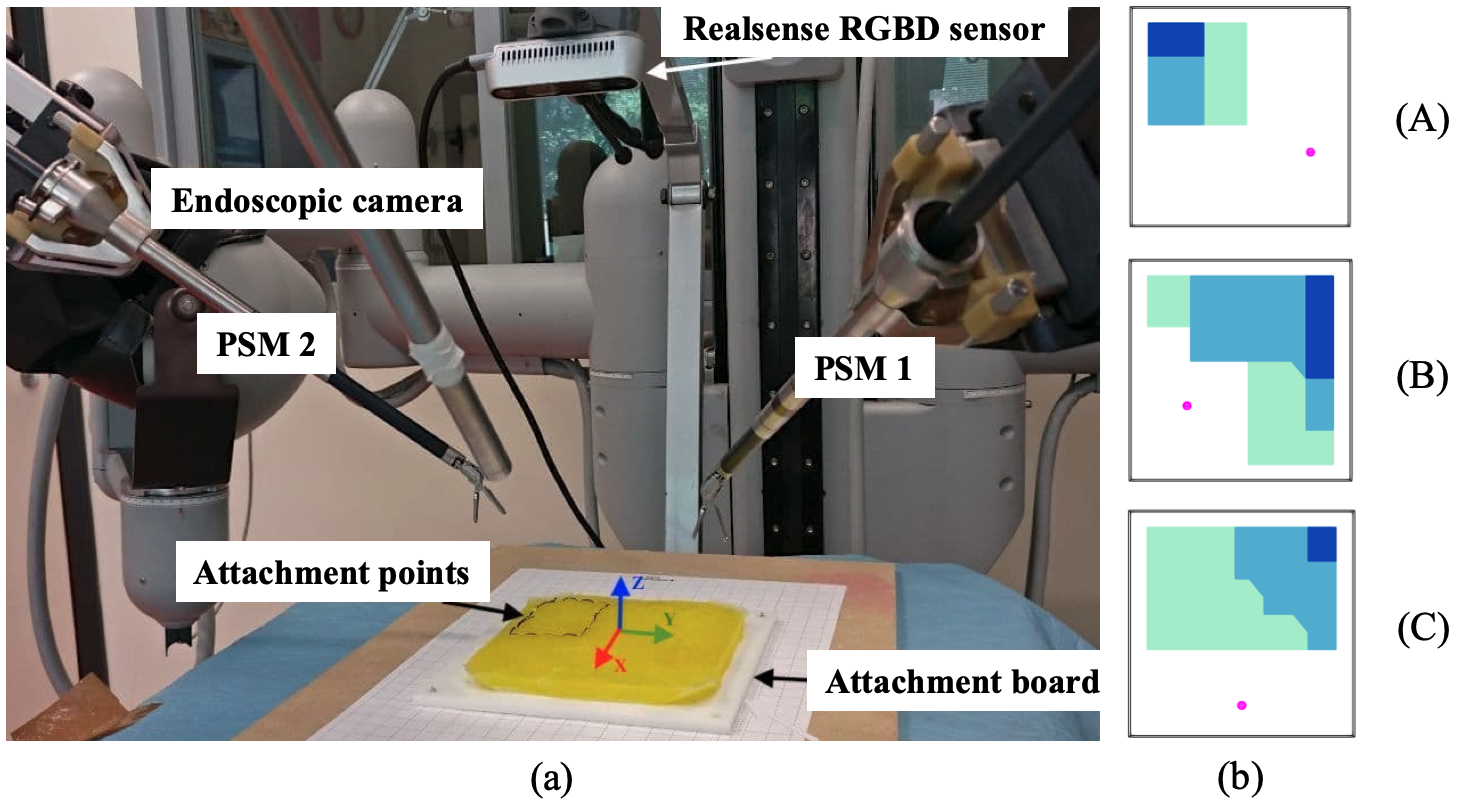}
\caption{(a) The setup for TR with dVRK. (b) Configurations of APs used for the real world experiments. The light blue region in each subimage represents the configuration of APs in the real setup ($\oeq$). Blue and green regions indicate pre-operative configurations with fewer APs ($\ominus$) and more APs ($\oplus$) than the real configuration. Magenta sphere represents the ROI.}
\label{fig:config-real}
\end{figure}

The performance of DEFRAS is evaluated on autonomous TR of silicone tissue with the real dVRK (Fig.~\ref{fig:config-real}). 
Similarly to \cite{tagliabue2021data}, APs are defined by stitching the tissue on an attachment board with regularly spaced holes every 10\,mm.
The point cloud of the tissue in real environment is acquired after color-based segmentation and decimation of the raw point cloud from Realsense as in \cite{tagliabue2021intra}, and compared with the point cloud from the simulated environment for monitoring.
Visibility in the real environment is computed detecting the ROI (with known size and position) from the endoscopic camera images with color-based segmentation and considering the percentage of visible pixels, as in \cite{pore2021learning}.

We consider three real world scenarios (A, B and C), obtained by stitching the tissue to the attachment board with 3 different configurations of APs (light blue in Fig.~\ref{fig:config-real}b). For each scenario, we consider 3 pre-operative models, each characterized by a different pre-operative knowledge of APs: (1) fewer APs (between 20-60\%), (2) more APs (between 140-160\%), and (3) same APs than/as the real scenario (Fig.~\ref{fig:config-real}b). For each of them, we generate the pre-operative plan in simulation, and test DEFRAS on the corresponding real configuration with the dVRK.  

Moreover, we analyze the performance of our framework with respect to two state-of-the-art implementations of autonomous TR on real setups. In particular, we select the work by Pore et al. \cite{pore2021learning}, which proposes reinforcement learning to generate an optimal motion-level policy for retraction, and the one by Attanasio et al. \cite{attanasio2020autonomous}, which relies on image-guided control to reveal a ROI behind tissue flap. Similarly to us, both these approaches evaluate task success as percentage of ROI exposure upon task completion.

\section{RESULTS AND DISCUSSION}
\label{sec:res}
\begin{table*}[t]
\caption{Simulation experiments with fewer ($\ominus$)
and more ($\oplus$) APs than pre-operative configuration. For each metric, upper row is referred to DEFRAS (\textit{D}), lower row to FRAS (\textit{F}). ``BANet" row reports the success rate of APs update with BANet in DEFRAS.}
\label{tab:results_sim}
\centering
\begin{tabular}{|c|m{.2cm}|m{.3cm}m{.3cm}|m{.3cm}m{.3cm}|m{.3cm}m{.3cm}|m{.3cm}m{.3cm}|m{.3cm}m{.3cm}|m{.3cm}m{.3cm}|m{.3cm}m{.3cm}|m{.3cm}m{.3cm}|cc|}

& & \multicolumn{2}{c}{\textbf{0}}& \multicolumn{2}{c}{\textbf{1}}& \multicolumn{2}{c}{\textbf{2}}& \multicolumn{2}{c}{\textbf{3}}& \multicolumn{2}{c}{\textbf{4}}& \multicolumn{2}{c}{\textbf{5}}& \multicolumn{2}{c}{\textbf{6}}& \multicolumn{2}{c|}{\textbf{7}} & & \\ 
 & & $\ominus$ & $\oplus$ & $\ominus$ & $\oplus$ & $\ominus$ & $\oplus$ & $\ominus$ & $\oplus$ & $\ominus$ & $\oplus$ & $\ominus$ & $\oplus$ & $\ominus$ & $\oplus$ & $\ominus$ & $\oplus$ & Median & IQR \\ 
\toprule 
multirow{2}{*}{Visibility [\%]} & \textit{D} & 100 & 100 & 100 & 100 & 100 & 0 & 9 & 95 & 100 & 100 & 0 & 100 & 100 & 100 & 100 & 100 & 100 & 98-100\\ 
& \textit{F} & 100 & 90 & 100 & 0 & 0 & 0 & 90 & 0 & 100 & 100 & 100 & 0 & 97 & 93 & 93 & 93 & 93 & 0-100\\ 
\midrule 
\multirow{2}{*}{Actions} & \textit{D} & 4 & 11 & 6 & 6 & 6 & 30 & 38 & 12 & 5 & 26 & 29 & 25 & 5 & 30 & 5 & 5 & 8 & 5-26\\ 
& \textit{F} & 5 & 6 & 15 & 23 & 22 & 22 & 7 & 22 & 17 & 19 & 7 & 22 & 16 & 16 & 9 & 15 & 16 & 8-22\\
\midrule 
\multirow{2}{*}{$F>\epsilon$} & \textit{D} & 0 & 26 & 11 & 3 & 1 & 44 & 240 & 27 & 0 & 46 & 99 & 83 & 8 & 61 & 1 & 3 & 18 & 2-49\\ 
& \textit{F} & 0 & 0 & 29 & 33 & 32 & 28 & 11 & 40 & 21 & 29 & 0 & 28 & 27 & 18 & 20 & 12 & 24 & 11-29\\
\midrule 
\multirow{2}{*}{$F_{max}$ [N]} & \textit{D} & 0.2 & 1.3 & 0.4 & 0.4 & 0.4 & 1.4 & 0.9 & 1.8 & 0.3 & 1.7 & 3.3 & 3.3 & 0.4 & 1.2 & 0.3 & 0.4 & 0.6 & 0.4-1.5\\
& \textit{F} & 0.3 & 0.3 & 1.0 & 1.0 & 1.1 & 1.0 & 0.4 & 1.2 & 0.9 & 0.9 & 0.3 & 1.1 & 0.9 & 0.9 & 1.2 & 0.9 & 0.9 & 0.8-1.0\\
\midrule 
BANet & \textit{D} & 1/1 & 1/1 &  - & 0/1 & 2/2 & 1/1 & 6/6 &  - & 1/1 & 2/2 & 11/12 &  - & 1/1 & 4/4 & 2/2 &  - & - & - \\ 
\end{tabular}

\end{table*}

\begin{table*}[th]
\caption{Real world experiments with DEFRAS. For each real world configuration (A, B, C), we report details of TR executions when underestimating real APs ($\ominus$), overestimating real APs ($\oplus$) and having perfect knowledge of the real APs ($\oeq$).}
\label{tab:results_real}
\centering
\begin{tabular}{|c|c|c|c||c|c|c||c|c|c||c|c|}
& \multicolumn{3}{|c||}{\textbf{A}}& \multicolumn{3}{c||}{\textbf{B}}& \multicolumn{3}{c||}{\textbf{C}} & & \\ 
& $\ominus$ & $\oplus$ & $\oeq$ & $\ominus$ & $\oplus$ & $\oeq$ & $\ominus$ & $\oplus$ & $\oeq$ & Median & IQR \\ 
\toprule
Visibility [\%] & 99 & 95 & 100 & 100 & 100 & 71 & 99 & 83 & 96 & 99 & 95-100\\ 
Actions & 13 & 15 & 9 & 24 & 7 & 8 & 14 & 7 & 7 & 9 & 7-14\\ 
$F>\epsilon$ & 11 & 20 & 8 & 0 & 8 & 20 & 8 & 8 & 6 & 8 & 8-11\\ 
$F_{max}$ [N] & 1.2 & 1.1 & 0.5 & 0.2 & 0.5 & 1.1 & 1.1 & 0.7 & 0.5 & 0.7 & 0.5-1.1\\ 
BANet & 1/1 & 1/1 & 1/1 & 13/14 & - & - & 1/1 & - & - & - & - \\ 
\end{tabular}
\end{table*}

\subsection{Simulation Experiments}
The optimal plans for each pre-operative configuration successfully expose the ROI with a maximum of 6 actions, with the force exceeding the safety threshold $\epsilon$ at most once, but always keeping $F_{max}$ below \SI{0.4}{\N}.

Table~\ref{tab:results_sim} details the results obtained with DEFRAS and FRAS on scenarios with fewer and more APs than the corresponding pre-operative configuration. 
With DEFRAS, the ROI is successfully exposed in 13 out of 16 cases, even when the pre-operative information (i.e., knowledge of APs) is imprecise. 
Results show that, even when starting from pre-operative plans which minimize the tissue force, the force threshold can be exceeded during task execution, hence force feedback is needed to prevent dangerous tissue solicitation.
The tissue force reaches higher values ($>\SI{1}{\N}$) mostly in scenarios with more APs than the pre-operative configuration ($\oplus$ in Table). This is motivated by the fact that optimal plans might select a grasping point close to APs that are unknown pre-operatively. In this case, monitoring detects a discrepancy between the pre-operative simulation and the actual simulation, and triggers learning for APs update close to the current grasping point, resulting in high force estimate.
The results of our framework also show that when pre-operative APs overestimate actual APs ($\ominus$ in Table), fewer actions are needed with respect to the case when APs are underestimated.
Therefore, overestimating APs in the pre-operative phase leads to more optimized task executions, reducing (potentially unsafe) interaction with the tissue.

From the comparison with FRAS, the presence of the pre-operative plan in DEFRAS proves fundamental to optimize executions in terms of tissue solicitation.
Monitoring and learning modules in DEFRAS also improves performance. 
Whenever the mismatch between the pre-operative simulation and the actual simulation causes the monitoring module to trigger update of APs, BANet is able to provide an updated estimate of APs that brings the monitoring metric below threshold, except cases $1\oplus$ (task success, ROI is exposed) and $5\ominus$ (the task fails).
Overall, the integration of the proposed deliberative modules results in higher median visibility (100\% vs. 93\%), lower number of failed tasks (3 vs. 5) and less frequent overcoming of the force threshold (-25\% times, with -30\% median force) with respect to FRAS.
We notice that FRAS is able to solve the task when DEFRAS fails in configurations $3\ominus$ and $5\ominus$. However, in both cases the monitoring module triggers APs update and the force threshold is overcome more times than all other configurations. This suggests that the pre-operative knowledge of APs is particularly far from reality in these cases, meaning that the estimate of the force from the pre-operative simulation provided by FRAS might be unreliable.

\subsection{Real World Experiments}
Table \ref{tab:results_real} presents the results of experiments on the real setup. 
Obtained results confirm that whenever pre-operative APs underestimate real APs, the monitoring and learning modules are triggered more frequently. The update of APs leads to successful task completion in all cases. Moreover, when the simulated and real APs match, the number of actions is usually minimum (except for configuration B, with 8 actions for $\oeq$ and 7 for $\oplus$). This suggests that overall the optimal pre-operative plan leads to better performances when APs are perfectly known. 

In the video attachment, we show the executions in configuration A. The video shows that DEFRAS is able to re-plan and complete the task in case of grasping close to unknown APs ($\ominus$) and null visibility after optimal plan ($\oplus$). Re-planning proves able to successfully expose the ROI in both cases. Monitoring module detects a mismatch also in $\oeq$ case, due to uncertainty introduced by real cameras and the calibration procedure. However, DEFRAS is able to cope with such situation and successfully complete the task.

The average visibility achieved by DEFRAS is 94\%, which is aligned with the 90\% visibility obtained by \cite{pore2021learning}, and improves the one reported in \cite{attanasio2020autonomous} (83\%). 
Both reference works do not implement monitoring and learning.

\section{CONCLUSIONS}
\label{sec:conclusion}
In this paper we proposed DEFRAS, a deliberative framework for autonomous robotic surgery.
We introduced the fundamental capabilities of monitoring and learning in the context of a logic-based framework for autonomous interpretable planning, to deal with the uncertainty of anatomical environments. 
Experiments on a representative soft tissue retraction task have shown that our framework can learn and refine available pre-operative knowledge about APs in an online simulation, relying on constant monitoring of sensors and simulation data.
In this way, any pre-operative plan can be adapted, and tissue force information from the simulation remains reliable for task completion. Compared with a framework lacking deliberative functions and recent tissue retraction implementations, results show that the percentage of task success increases, especially when prior APs are overestimated. Moreover, reliance on a pre-operative patient-specific plan brings benefits in terms of minimal manipulation of the tissue, hence reduced risk of tissue damage.

The modularity of DEFRAS will allow us to improve individual deliberative capabilities in order to deal with more realistic and challenging surgical scenarios in future works. Motion adaptation and learning as in \cite{ginesi2021overcoming,pore2021learning} will be implemented to replicate expert surgeons’ dexterity.
Furthermore, we plan to improve task knowledge with pre-operative \cite{bombieri2021automatic} and intra-operative learning \cite{meli2021inductive,meli2020towards}.
Finally, we will cope with surgical-specific problems as vision occlusion.

\bibliographystyle{./IEEEtran}
\bibliography{./IEEEabrv,./bibliography}

\end{document}